\documentclass[10pt,twocolumn,letterpaper]{article}

\usepackage{cvpr}
\usepackage{times}
\usepackage{epsfig}
\usepackage{graphicx}
\usepackage{amsmath}
\usepackage{amssymb}

\usepackage{subfig}
\usepackage{caption}
\usepackage{multirow}
\usepackage{float}

\usepackage[pagebackref=true,breaklinks=true,colorlinks,bookmarks=false]{hyperref}

\cvprfinalcopy 

\def\cvprPaperID{9614} 

\ifcvprfinal\fi
\begin{document}

\title{ProAlignNet : Unsupervised Learning for Progressively Aligning Noisy Contours}

\author{VSR Veeravasarapu, Abhishek Goel, Deepak Mittal, Maneesh Singh\\
Verisk AI, Verisk Analytics \\
{ \tt\small \{s.veeravasarapu, a.goel, deepak.mittal, msingh\}@verisk.com }
}

\maketitle

\begin{abstract}
Contour shape alignment is a fundamental but challenging problem in computer vision, especially when the observations are partial, noisy, and largely misaligned. 
Recent ConvNet-based architectures that were proposed to align image structures tend to fail with contour representation of shapes, mostly due to the use of proximity-insensitive pixel-wise similarity measures as loss functions in their training processes. 
This work presents a novel ConvNet, ``\textit{ProAlignNet}", that accounts for large scale misalignments
and complex transformations between the contour shapes.
It infers the warp parameters in a multi-scale fashion with progressively increasing complex transformations over increasing scales.
It learns --without supervision-- to align contours, agnostic to noise and missing parts, by training with a novel loss function which is derived an upperbound of a proximity-sensitive and local shape-dependent similarity metric that uses classical Morphological \textit{Chamfer} Distance Transform.
We evaluate the reliability of these proposals on a simulated MNIST noisy contours dataset via some basic sanity check experiments. Next, we demonstrate the effectiveness of the proposed models in two real-world applications of (i) aligning geo-parcel data to aerial image maps and (ii) refining coarsely annotated segmentation labels. In both applications, the proposed models consistently perform superior to state-of-the-art methods. 
\vspace{-15pt}
\end{abstract}

\section{Introduction}
\begin{figure}[t!]
    \centering
    \setlength\tabcolsep{0.1pt}
    \begin{tabular}{c}
        Source \\
        \includegraphics[width=0.11\textwidth]{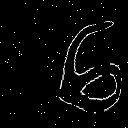}\\
        \includegraphics[width=0.11\textwidth]{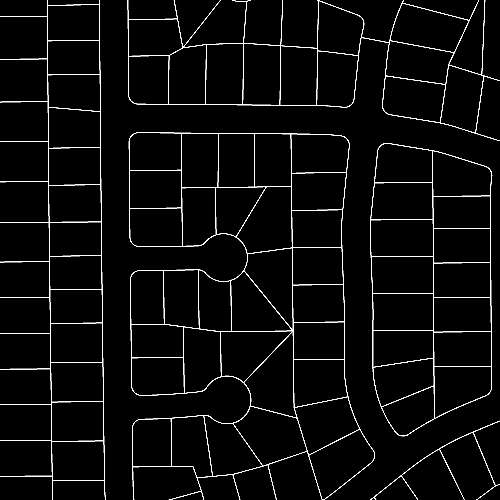}\\
        \includegraphics[width=0.11\textwidth]{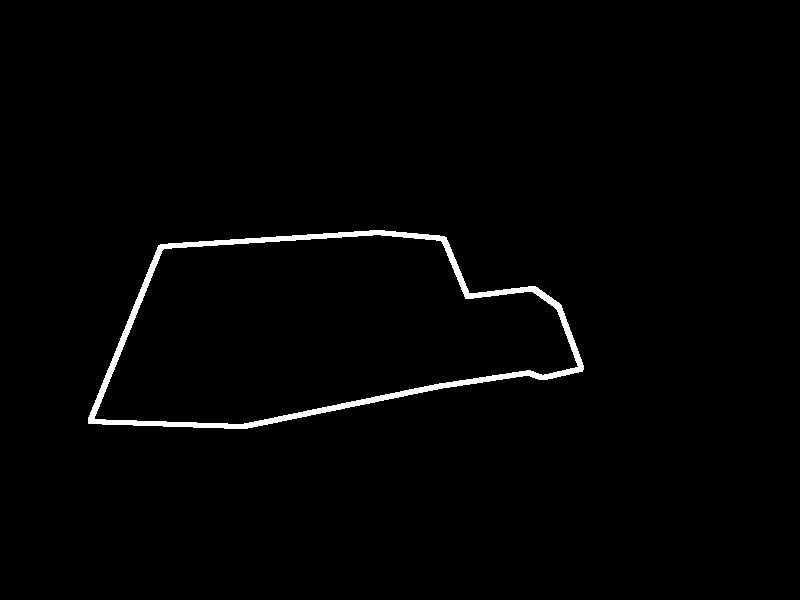}
    \end{tabular}
    \begin{tabular}{c}
        Target\\
        \includegraphics[width=0.11\textwidth]{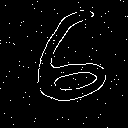}\\
        \includegraphics[width=0.11\textwidth]{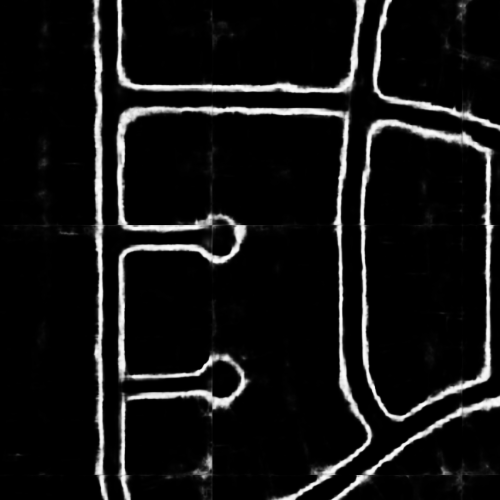}\\
        \includegraphics[width=0.11\textwidth]{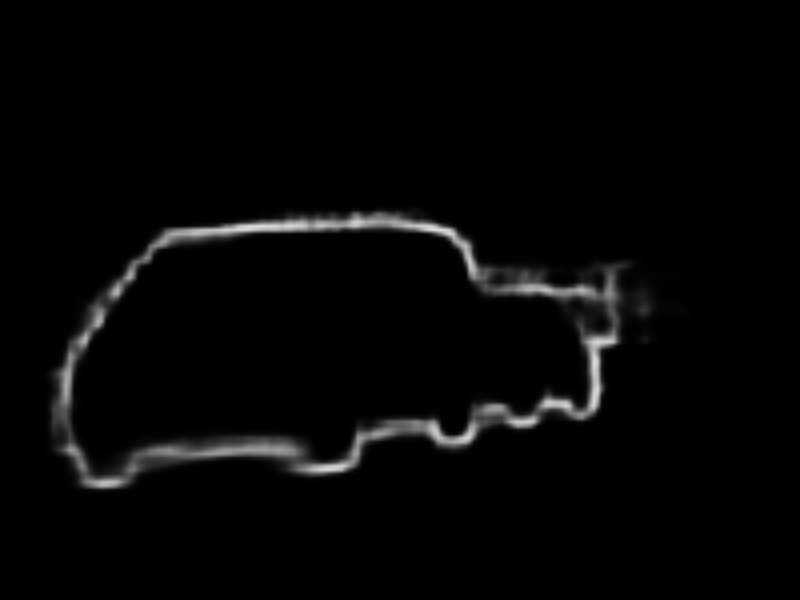}
    \end{tabular}
    \begin{tabular}{c}
        Aligned\\
        \includegraphics[width=0.11\textwidth]{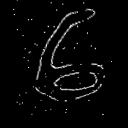}\\
        \includegraphics[width=0.11\textwidth]{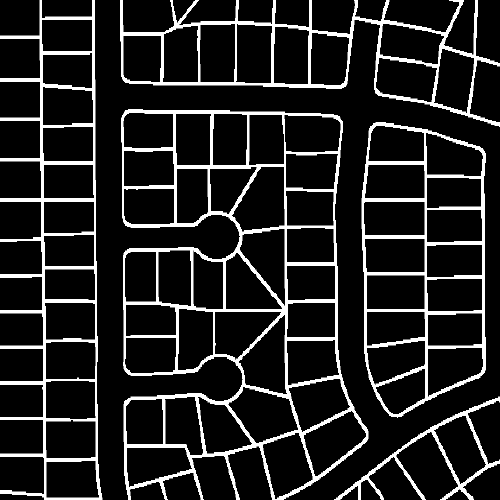} \\
        \includegraphics[width=0.11\textwidth]{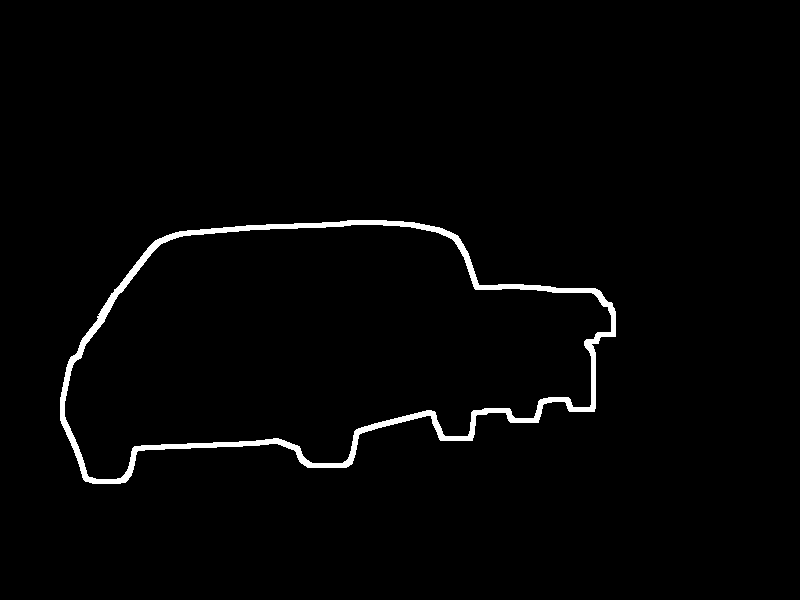}
    \end{tabular}
    \begin{tabular}{c}
        Overlaid\\
        \includegraphics[width=0.11\textwidth]{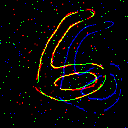} \\
        \includegraphics[width=0.11\textwidth]{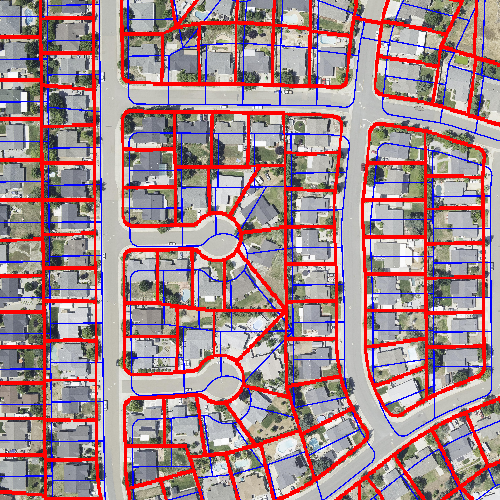} \\
        \includegraphics[width=0.11\textwidth]{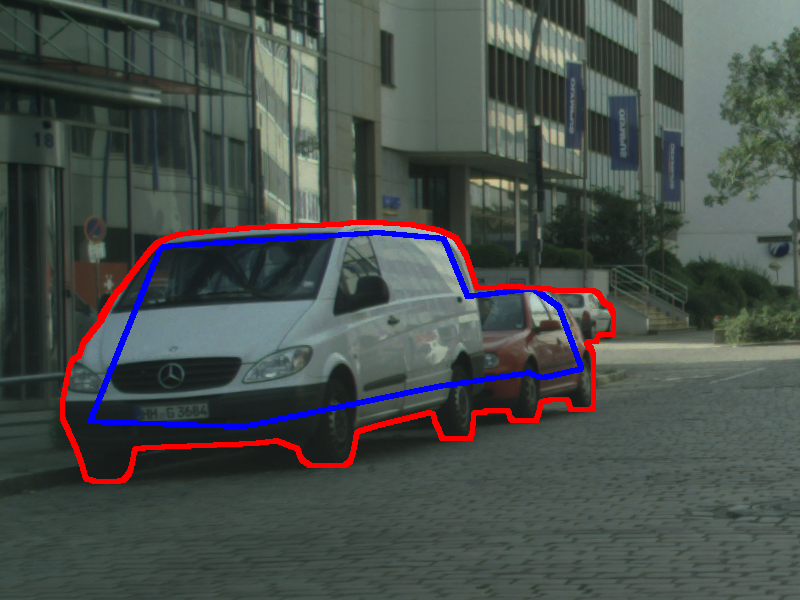}
    \end{tabular}
    \caption{\small This work considers the problem of learning to align source (1st column) with target (2nd column) contour images. Aligned results are shown in 3rd column. Visualizations (4th column) are input image canvases overlaid with original (Blue) and aligned (Red) source contours. Three rows contain sample results from the applications we considered in this work: (i) noisy digit contour alignment (ii) geo-parcel alignment and (ii) coarse-label refinement.}
    \label{fig_intro}
    \vspace{-15pt}
\end{figure}

Contour shape alignment with noisy image observations is a fundamental, but challenging, problem in computer vision and graphics fields, with diverse applications including skeleton/silhouette alignment \cite{alexiadis2011evaluating} (for animation re-targeting), semantic boundary alignment \cite{yu2018simultaneous} and shape-to-scan alignment \cite{hirshberg2012coregistration} etc. 
For instance,  consider the first row of Figure \ref{fig_intro}. It represents a process of geo-parcel alignment that requires aligning geo-parcel data (legal land boundaries maintained by local counties) to aerial image maps. 
These two modalities of geo-spatial data, if well aligned, are useful to assist the processes of property assessment and tax/insurance underwritings. 
Classically, contour alignment problems have been approached by finding key points or features of the shapes and aligning them by optimizing for the parameters of a predefined class of transformations such as affine or rigid transforms \cite{saxena2014survey}. These methods may not work for the shapes whose alignment requires a transform different from the hired ones. Nonrigid registration methods have also been proposed in the literature, mainly using intensity-based similarity metrics \cite{klein2009elastix}. However, these methods are often computationally expensive and sensitive towards corrupted parts of the shapes.

Motivated by the strong invasion and great success of deep convolutional neural networks (ConvNets) in various vision tasks, some recent works \cite{kanazawa2016warpnet,miao2018dilated,guan2019deformable,hanocka2018alignet,de2017end} have designed ConvNet based architectures for shape alignment/registration and shown impressive results on the datasets with limited misalignments. However, we observe that these approaches tend to fail in noisy, partially observed and largely misaligned contour shape contexts. We believe that this is due to (a) direct prediction of alignment warp in one-go at input resolution and (b) training with proximity-insensitive pixel-level similarity metrics (such as normalized cross correlation \cite{de2017end}) that might result in non-informative gradients as the spatial overlap between contours may not be significant. 

To overcome the problems mentioned above, we propose a novel ConvNet architecture, ``\textbf{\textit{ProAlignNet}}" that learns to align a pair of contour images while being robust to noise and partial occlusions in the images. It finds an optimal transformation that aligns source to target contours in a multi-scale fashion with progressively increasing complex transformations over increasing scales. 

We also propose a novel loss function that accounts for not only local shape similarity but also proximity of similar shapes. The proposed loss is based on classical \textit{Chamfer} distance that measures the proximity between the contours, thus, provides informative gradients even though there is no spatial overlap between the contours.  Chamfer distance was a popular metric \cite{borgefors1984distance,gavrila1998multi} for binary shape alignment and  was traditionally implemented efficiently using morphological distance transforms \cite{butt1998optimum}.
However, this morphological chamfer distance transform (MCDT) is nondifferentiable wrt the warp parameters, which makes it nontrivial to use with BackProp. Hence, we device a reparameterization trick (inspired from homeomorphism properties of the transformation families we use) to make it usable with BackProp. 
However, this trick requires a backward/inverse transform (that aligns target to source) to be measured. Here, we propose to swap the roles of source and target features in the alignment modules to get the backward transform with the same network components without any additional parameters/layers. As a side benefit, this forward-backward transform consistency constraint acts as a powerful regularizer that improves generalization capabilities. 

Chamfer distance is amenable to noise as it uses Euclidean distance to find nearest neighbours, and it neglects local shape statistics around pixels. Hence, we introduce a data-dependent term into distance function that MCDT uses. This modification increases the proposed loss's robustness to noise pixels. We then derive an upperbound of this loss function, which is differentiable and computationally efficient.  Since we use one instance of the loss at each scale, this multiscale objective passes gradients to all the layers simultaneously. It helps the training process to be more stable and less sensitive to hyperparameters such as learning rates as reported in  \cite{karnewar2019msg}. Our training process is completely unsupervised as it does not require any ground-truth (GT) warps that the network is expected to reproduce. By training with a loss function that accounts for local shape contexts while being invariant to corrupted pixels, our network learns to be robust to noise and partial occlusions.

\begin{figure*}[ht!]
    \centering
    \includegraphics[width=\textwidth, height=7cm]{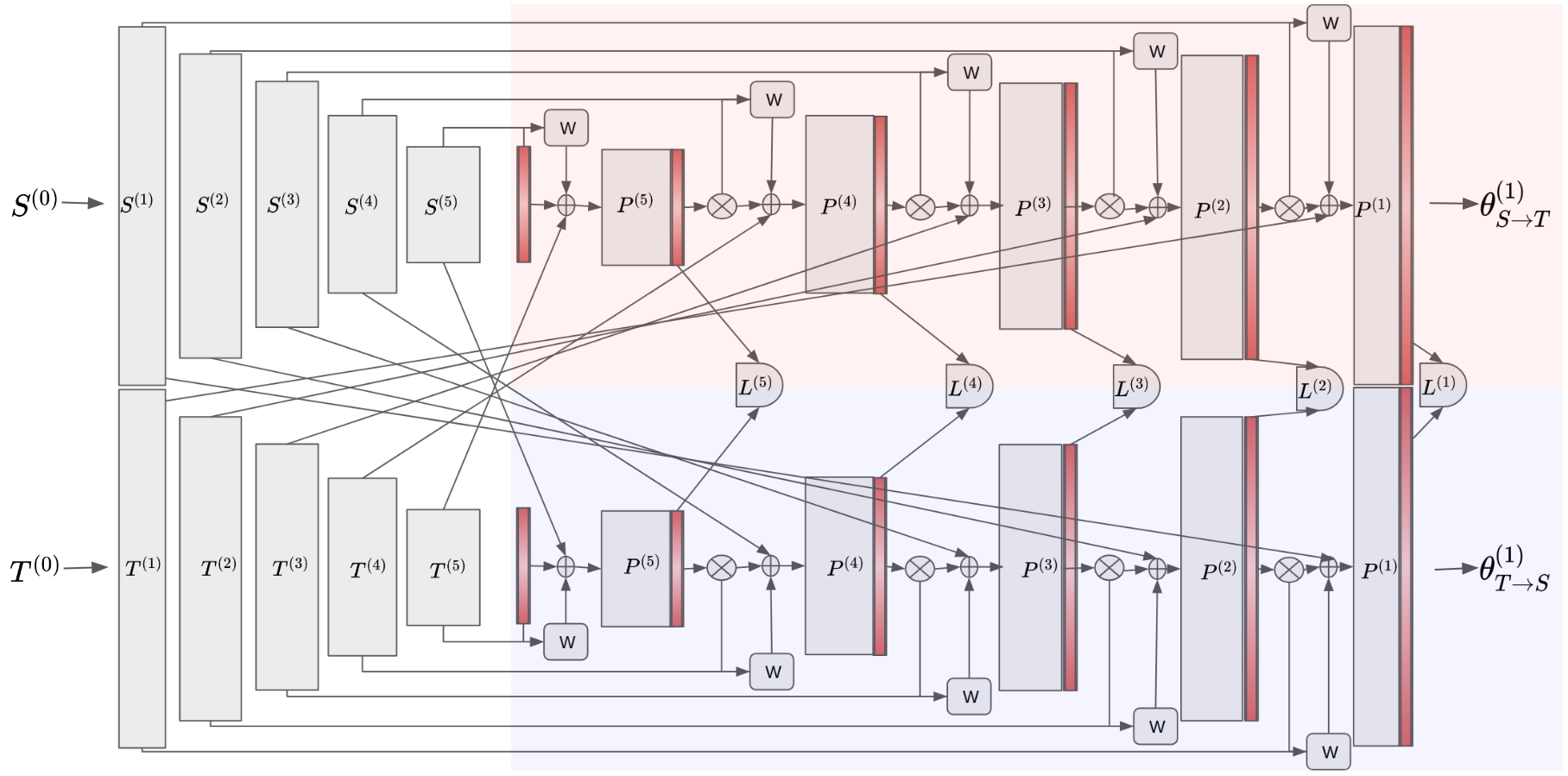}
    \caption{\small Overview of ProAlignNet with 5 Scales: Given a pair of source ($S^{(0)}$) and target ($T^{(0)}$) contour images, we first use base CNNs to get the feature maps.  A cascade of Warp predictors ($P^{(i)}$) operates in a multiscale fashion to predict transformations to align source to target features. Coarsest predictor starts with identity transformation and uses affine transform to refine it.  Complexity of transformation used in these predictors progressively increases over scales. Finest scale predictor used more complex transformation such as thin-plate-splines with higher number of control points. }
    \label{fig_pipeline}
    \vspace{-15pt}
\end{figure*}

In summary, our contributions are the following:
\begin{itemize}
\setlength\itemsep{0.05pt}
 \item \textbf{ProAlignNet}: a novel multiscale contour alignment network that employs progressively increasing complex transforms over increasing scales.
    \item \textbf{Shape-sensitive Chamfer Upperbound Loss}: a novel loss function that measures proximity and local shape similarity while being robust towards noise and partial occlusions.
    \item Ablation studies with \textbf{MNIST noisy digit contours}. 
    \item Demonstration of efficacy of the proposed models in two real-world applications: \textbf{geo-parcel alignment} and \textbf{coarse-to-fine label refinement}. 
\end{itemize}

\section{Motivations and Background}

\textbf{Deep Learning for Shape Alignment}: Several works have employed deep networks \cite{kanazawa2016warpnet,miao2018dilated,guan2019deformable} to directly estimate warp parameters and trained using ground-truth warp fields/parameters. However, it is challenging to collect ground-truth warp fields for several real-world applications, especially for nonrigid alignment scenarios. Hence, recent works \cite{hanocka2018alignet,de2017end} propose unsupervised methods for nonrigid registration and these are closest to our approach. The work of \cite{de2017end} designed a deep network referred to as DIRNet, consists of a CNN based regressor that predicts a deformation field, followed a spatial transformation function that warps source to target image. The regressor takes concatenated pairs of source and target images and predicts displacement vector fields directly from convolutional features. These fields are then upsampled using splines to original resolutions and used to warp the source images. These models were trained with pixel-wise similarity metrics; thus, they can deal only with small scale deformations. Similarly the work of \cite{hanocka2018alignet} proposed a ConvNet referred as AlignNet which constitutes a regression network and a novel integrator layer that outputs free-form deformation field. We empirically show that these approaches might fail in the contexts of contour shapes with large misalignments. Unlike these methods, our ProAlignNet uses inference process that accounts for large scale misalignments and complex transformations between the contour shapes by inferring in a multi-scale fashion with progressively increasing complex transformations over increasing scales.

\textbf{Chamfer distance}: Proximity metrics such as Chamfer distances and Earth-mover distances \cite{grauman2004fast} are more popular in shape correspondences. 
They are criticized for their computationally cost. However, when shapes are represented in binary contour images, morphological distance transforms \cite{nacken1994chamfer} can be used to compute the Chamfer distance between two contour images efficiently. However, it has two problems in its vanilla form: (i) nondifferentiability: morphological distance transform has a process of collecting all nonzero pixels to find the nearest one. This set-collection operation is nondifferentiable; (ii) sensitivity towards noise: while computing distance to the nearest pixel, it is blind to noise pixels. This influences the distance estimate between shapes. In this work, we derive a loss function, which is based on Chamfer distance but overcomes problems mentioned above.

\section{ProAlignNet} \label{sec_approach}
The network architecture of the \textit{ProAlignNet}, as shown in Figure \ref{fig_pipeline}, consists of a set of simple modules: 
(i) Base CNN blocks to extract features ($S^{(i)}$ and $T^{(i)}$) of source and target images at different scales; 
(ii) Warp predictors ($P^{(i)}$) that predict warp fields with a predefined class of transformations at each scale; 
(iii) Warp layer ($W$) that warps the features using warp fields. 
Here, $i$ represents the index of scale with $i=0$ and $i=K$ denoting the original (finest) and coarsest scales respectively. 

\subsection{Base CNNs for Feature Extraction}
Given the source and target input image pair $(S^{(0)}, T^{(0)}) \in R^{h\times w\times c}$, we extract their feature representations $\big\{ \big(S^{(i)}, T^{(i)}\big)_{i=1}^K \big\}$ from different layers $i \in \{1,2,...,K\}$ of fully-convolutional backbone networks $F_s$ and $F_t$. As shown in Figure \ref{fig_pipeline}, we use $K=5$ in all experiments provided in this work. 

\subsection{Warp Predictors and Warp Layer}
Once we extract the features from both source and target, we design a cascade of warp predictors ($P^{(i)}$) that learn to predict transformations to align source to target features at multiple scales. Each $P^{(i)}$ block takes a feature tensor that is a result of concatenating (denoted by $\oplus$ in Figure \ref{fig_pipeline}) three features: (i) source features warped by the warp field from previous coarser scale but upsampled (denoted by $\otimes$) by a factor of 2, (ii) target features, and (iii) upsampled warp field from previous coarser scale. Mathematically, warp predictor at a given scale $i$ learns the below functionality, 
\begin{equation*}
    P^{(i)}: S^{(i)}(\theta^{i+1}_{\otimes2}) \oplus T^{(i)} \oplus \theta^{i+1}_{\otimes2} \longrightarrow \theta^{(i)}
\end{equation*}
where $\theta_{\otimes u}$ denotes warp field $\theta$ upsampled by a factor of $u$. 
We initialize $\theta^{K+1}$ as identity warp field at coarsest scale $K$. 

A warp or transformer layer ($W$) uses the predicted warp parameters to derive displacement field and warp the source features. 
These warp predictors by design resemble Spatial Transformer Networks proposed in \cite{jaderberg2015spatial}.  
In general, they have a regression network that predicts parameters of the transformations, followed by a grid layer that converts parameters to pixel-level warp field. 

In Figure \ref{fig_pipeline}, the red shaded part of ProAlignNet estimates a forward transformation that aligns source to target. It initializes the transformation with identity warp grid and refines it using multiscale warp predictors. It starts the refinement process by using simple affine transformation at the coarsest scale, which is a linear transformation with 6 DOF. The estimated affine transform grid is used to align source features using warp layer. The aligned features are then passed along with estimated affine grid through the next finer scale network, which estimates a warp field using more flexible transformations such as thin-plate splines(\textit{tps}).  As we move toward the finest scale, we use \textit{tps} transforms with increasing resolutions of control point grid. The final estimate of the geometric transformation is then obtained from the finest scale predictor that learns to compose all coarser transformations along with local refinements that align its input features. As we explain in the next section, our loss function requires backward transform also to be measured that aligns target to source. This backward transform is estimated using the same warp predictor components but with a second pass (blue shaded part in Figure \ref{fig_pipeline}) by reversing the roles of source and target features. 

\subsection{Multi Scale Loss Objective}
We use one instance of the loss function, $L^{(i)}$, at each scale $i$ as shown in Figure \ref{fig_pipeline}. 
Hence, the multiscale objective we use to train ProAlignNet is given as, 
\begin{equation}
    L_{MS} = \sum_{i=1}^K \lambda_i L^{(i)} = \sum_{i=1}^K \lambda_i L\bigg(S^{(0)}\big(\theta^{(i)}_{\otimes 2^i}\big), T^{(0)}\bigg)
\end{equation}
$\theta^{(i)}_{\otimes 2^i}$ denotes the warp field predicted at scale $i$ but upsampled by a factor $2^i$ that brings up the field to be at similar resolution as the input source and target images $S^{(0)}$ and $T^{(0)}$. $\lambda_i$ is the weighting factor for the loss $L$ from scale $i$. $L$ can be any alignment loss function, for instance NCC or MSE. However, motivated by the fact that pixel-wise similarity metrics suffer in large scale misalignment by not considering the proximity of shapes we propose a novel loss function in this work. 

\section{Shape-dependent Chamfer Upperbound} \label{sec_losses}
\subsection{Chamfer Loss}
Chamfer distance was a popular metric to measure the proximity between two shapes or contours. 
It was initially proposed for point sets. The chamfer distance b/w any two point sets $X$ and $Y$ is given as, 
\begin{equation}\label{eq_sym_chamfer}
    C(X, Y) = \frac{1}{N_X} \sum_{x\in X} \min_{{y\in Y}}E(x,y) + \frac{1}{N_Y} \sum_{y\in Y} \min_{{x\in X}} E(y,x)
\end{equation}
where $E(x,y)$ is Euclidean distance between the points $x$ and $y$. $N_X$ and $N_Y$ denotes the cardinality of the sets $X$ and $Y$ respectively.  
In the binary image representation of shapes, one can use the concepts of morphological distance transform to efficiently compute the Chamfer distance between two images. Morphological Distance Transform (MDT) computes Euclidean distance to nearest nonzero neighbor for each pixel $x$ in a given contour image $I$. Thus, it is represented by $dt[I](x) = \min_{i \in I} E(x,i)$. Using MDT, Chamfer distance between the source ($S$) and target ($T$) images can be written\footnote{Here dot (.) represents a scalar product.} as $ C(S, T) =  \frac{1}{N_S} dt[S].T + \frac{1}{N_T} S.dt[T]$. Here, $N_S$ and $N_T$ denotes number of nonzero pixels in $S$ and $T$ respectively. There are several efficient implementations available to compute $dt[.]$ (in Opencv, Scikit-image etc.). 
Now the chamfer loss b/w warped source and target at any scale (we drop scale $i$ for simplicity) becomes, 
\begin{equation} \label{chamfer_mdt}
    C\big(S(\theta), T\big) = \frac{1}{N_S} dt[S(\theta)].T + \frac{1}{N_T} S(\theta).dt[T]
\end{equation}
The gradient of the above loss function requires the distance transform operation $dt$ to be differentiable wrt warped source ($S(\theta)$). Unfortunately, $dt$ is not differentiable as it has a set-collection process to collect all nonzero pixels.  

\subsection{Reparametrized Chamfer Loss}
However, we overcome this problem by using a reparameterization trick inspired by homeomorphism \cite{moore2007evolution} properties of affine/\textit{tps} transformations. 
This property states that if a forward transformation $\theta_{S\rightarrow T} \in \Theta$ aligns $S$ with $T$ then there exists a $\theta_{T\rightarrow S}$ also $\in \Theta$ that aligns $T$ with $S$, given that $\Theta$ is a homeomorphic transformation group. This results in a corollary that $dt[S(\theta_{S\rightarrow T})].T = dt[S].T(\theta_{T\rightarrow S})$
(Please refer to  Supplementary for the proofs). 
With this reformulation, Eq \ref{chamfer_mdt} becomes
\begin{equation} \label{eq_bidirectional_loss}
    C_r\big(S(\theta), T\big) = dt[S].T(\theta_{T\rightarrow S}) + S(\theta_{S\rightarrow T}).dt[T]
\end{equation}

The gradient of the above loss function doesn't require $dt$ to be differentiable. Distance transform ($dt$) maps can be computed externally and supplied as reference signals. We refer to this loss as Reparametrized Chamfer loss. However, the above loss requires backward transform $\theta_{T \rightarrow S}$ to be estimated. 

\textbf{Bi-directional Transform Consistency}: For affine, one can analytically compute backward transform ($\theta_{T\rightarrow S}$) using matrix inverse. However, it is not trivial to get analytical inverse for \textit{tps} and other fully flexible transforms. Hence, we propose a simple and effective way to get backward transforms by a second pass through warp predictors by swapping the roles of source and target features. The loss in Eq \ref{eq_bidirectional_loss} constraints these forward and backward transforms to be consistent with each other. This constraint acts as a powerful regularizer on the network training. In Section \ref{exp_mnist}, we empirically demonstrate that it improves the generalization capabilities of the models. 

\subsection{Shape-dependent Chamfer Upperbound}
\textbf{Local-shape dependency}: 
The above Chamfer loss (Eq \ref{eq_bidirectional_loss}) is susceptible to noise and occlusions in the contour images as it computes the distance from nearest nonzero neighbor without checking if it is a noise pixel or indeed a part of contour.  To make it robust, we incorporate local shape dependency into the distance computation to make distance transform ($dt$) to choose nearest pixels based on not only spatial proximity but also local shape similarity.  Although several sophisticated local shape metrics are available, we stick to first-order intensity gradients \cite{sun2008learning,brox2004high} in this work for computational simplicity. More specifically, we consider unit gradients as a representation of local orientations of the contour pixels. We leave advanced local shape metrics for future work.
Here, we use a combination of Euclidean distances in Cartesian and image gradient space. 
Now local shape-dependent Chamfer distance is given by,  
\begin{equation}\label{eq_data_dependence}
\small
\begin{split}
    C_d(X, Y) = & \frac{1}{N_X} \sum_{x\in X} \min_{{y\in Y}} \bigg( E(x,y) + \alpha E(I_x', I_y') \bigg) \\ 
    & + \frac{1}{N_Y} \sum_{y\in Y} \min_{{x\in X}} \bigg( E(y,x) + \alpha E(I_y', I_x') \bigg)
\end{split}
\end{equation}
where $I_x'$ denotes the unit gradient vector computed at $x$. \\
\textbf{Upperbound of shape-dependent Chamfer loss}: 
However, we can not use the concept of MDT here as in Eq \ref{eq_bidirectional_loss} as the distance has a data-dependent term. Fortunately, $min$ arguments before distance terms in Eq \ref{eq_data_dependence} allow us to use a famous math property known as ``\textit{min-max inequality}." It results in an upperbound for Eq \ref{eq_data_dependence} with two simple terms.

We use min-max inequality to reformulate Eq \ref{eq_data_dependence} so that one can use the concepts of MDT and reparameterization as in Eq \ref{eq_bidirectional_loss}. Min-max inequality states that minimum of the sum of any two arbitrary functions $f(x)$ and $g(x)$ is upper-bounded by the sum of minimum and maximum of individual functions, i.e., $\min_x\big(f(x)+g(x)\big) \le \min_x f(x) + \max_x g(x)$ (Please refer to Supplementary for the proofs). 


Using the inequality for both terms on RHS of Eq \ref{eq_data_dependence} results an upperbound with original Chamfer distance (Eq 3) and shape-dependent term as follows, 
\begin{equation}
\small
\begin{split}
& C_d(X,Y) \le   C(X,Y) \\
& + \alpha \bigg( \frac{1}{N_X} \sum_{x \in X}\max_{y \in Y} E(I_x', I_y') + \frac{1}{N_Y} \sum_{y \in Y}\max_{x \in X} E(I_y', I_x')  \bigg) 
\end{split}
\end{equation}
Rewriting the above upperbound in the current context of warped source to target alignment,  
\begin{equation}
\small
\begin{split}
& C_d(S(\theta),T) \le   C(S(\theta),T) \\
& + \alpha \bigg( \frac{1}{N_{S(\theta)}} \sum_{x \in S(\theta)}\max_{y \in T} E(I_x', I_y') + \frac{1}{N_T} \sum_{y \in T}\max_{x \in S(\theta)} E(I_y', I_x')  \bigg)  
\end{split}
\end{equation}
We denote this upperbound as $C_{up}$. As one can observe, the shape-dependent terms are computationally heavy as the maximum being taken over the window of the entire image for each pixel in the other image. However, we can constraint this window to be local and search in the neighborhood defined by that window. Moreover, this maximum-finding operation can be implemented with \textit{MaxPool} layers. Finally, this local shape-dependent Chamfer upperbound is given by, 
\begin{equation}
\small
\begin{split}
    C_{up}(S, T) = & \bigg( \frac{1}{N_S} dt[S].T(\theta_{T\rightarrow S}) + \frac{1}{N_T} S(\theta_{S\rightarrow T}).dt[T] \bigg) + \\ 
    & \alpha \bigg( \frac{1}{N_{S(\theta)}}\sum_{x\in S(\theta)} \max_{{y\in T_x}} E(I_x',I_y')   + \frac{1}{N_T}\sum_{y\in T} \max_{{x\in S_y(\theta)}} E(I_y',I_x')\bigg)
\end{split}
\end{equation}




When the local window is restricted to be $1\times1$, minimizing the above term can be related to maximizing cross-correlation in intensity gradient space. When raw pixel intensities are used in place of gradients, this is maximizing NCC-related metric.  
Now the upperbound loss with unit gradients as local shape measures, 
\begin{equation} \label{eq_chamfer_upperbound}
\small
\begin{split}
    C_{up}\big(S(\theta), T\big) = & \bigg( \frac{1}{N_S}dt[S].T(\theta_{T\rightarrow S}) + \frac{1}{N_T}S(\theta_{S\rightarrow T}).dt[T] \bigg) \\
    &+ \alpha \bigg(   \frac{1}{N_{S(\theta)}} \sum_{x\in S(\theta)} \max_{{y\in T_x}} \sqrt{1-I_x'^T.I_y'} \\ 
    &+ \frac{1}{N_T} \sum_{y\in T} \max_{{x\in S_x(\theta)}} \sqrt{1-I_y'^T.I_x'}\bigg)
\end{split}
\end{equation}
Please refer to Supplementary for detailed derivations of the above equations.





\begin{figure}[ht!]
    \centering
    \subfloat[Source]{
    \setlength\tabcolsep{0.1pt}
    \begin{tabular}{c}
        \includegraphics[width=0.11\textwidth]{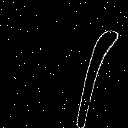} \\
        \includegraphics[width=0.11\textwidth]{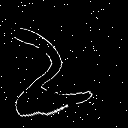} \\
        \includegraphics[width=0.11\textwidth]{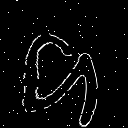} 
    \end{tabular}
     }
     \subfloat[Target]{
    \setlength\tabcolsep{0.1pt}
    \begin{tabular}{c}
        \includegraphics[width=0.11\textwidth]{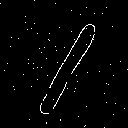} \\
        \includegraphics[width=0.11\textwidth]{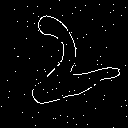} \\
        \includegraphics[width=0.11\textwidth]{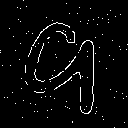} 
    \end{tabular}
     }
     \subfloat[Aligned]{
    \setlength\tabcolsep{0.1pt}
    \begin{tabular}{c}
        \includegraphics[width=0.11\textwidth]{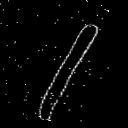} \\
        \includegraphics[width=0.11\textwidth]{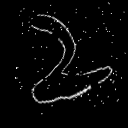} \\
        \includegraphics[width=0.11\textwidth]{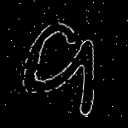} 
    \end{tabular}
     }
     \subfloat[Overlaid]{
    \setlength\tabcolsep{0.1pt}
    \begin{tabular}{c}
        \includegraphics[width=0.11\textwidth]{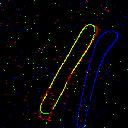} \\
        \includegraphics[width=0.11\textwidth]{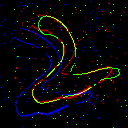} \\
        \includegraphics[width=0.11\textwidth]{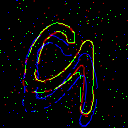} 
    \end{tabular}
     }
    \caption{\small Sample results on contourMNIST:  First, second and third column images are source, target and aligned source contour images respectively. Right most are the images prepared for better visualization with B,G,R channels as source, target, aligned images respectively.}
    \label{fig_mnist}
    \vspace{-15pt}
\end{figure}

\begin{table*}[ht!]
\centering
\small
\begin{tabular}{|l|l|l|c|l|}
\hline
Application                           & Method                                               & Loss                                     & Error ($\%px\le Z$)                    & Metric                                            \\ \hline
\multirow{6}{*}{contourMNIST}         & Given test pairs                                     & -                                        & 10.20 (39\%)                           & \multirow{6}{*}{\begin{tabular}[c]{@{}l@{}}Chamfer \\ Score: \\ lower the \\ better.\end{tabular}} \\
                                      & DIRNet   \cite{de2017end}           & Normalized Cross-Correlation             & 8.10 (46\%)                            &                                                   \\
                                      & DIRNet   \cite{de2017end}           & Local Shape-dependent Chamfer Upperbound & 4.83  (69\%)                           &                                                   \\
                                      & ALIGNet  \cite{hanocka2018alignet}  & Mean Squared Error                       & 6.69 (55\%)                            &                                                   \\
                                      & ALIGNet   \cite{hanocka2018alignet} & Local Shape-dependent Chamfer Upperbound & 4.04  (71\%)                           &                                                   \\
                                      & ProAlignNet                                          & Normalized Cross-Correlation             & 5.46 (64\%)                            &                                                   \\
                                      & ProAlignNet                                          & Asymmetric Chamfer                       & 3.38 (89\%)                            &                                                   \\
                                      & ProAlignNet                                          & Reparametrized bi-directional Chamfer    & 2.91 (93\%)                            &                                                   \\
                                      & ProAlignNet                                          & Local Shape-dependent Chamfer Upperbound & \textbf{2.17 (96\%)}  &                                                   \\ \hline
\multirow{4}{*}{\begin{tabular}[c]{@{}l@{}}Geo-parcel \\ alignment\end{tabular}} & Given test pairs                                     & -                                        & 42.38 (48\%)                           & \multirow{6}{*}{\begin{tabular}[c]{@{}l@{}}Chamfer \\ Score: \\ lower the \\ better.\end{tabular}} \\
                                      & DIRNet \cite{de2017end}             & Normalized Cross-Correlation             & 41.36 (48\%)                           &                                                   \\
                                      & DIRNet \cite{de2017end}             & Local Shape-dependent Chamfer Upperbound & 27.70 (65\%)                           &                                                   \\
                                      & ALIGNet \cite{hanocka2018alignet}   & Mean Squared Error                       & 39.98 (59\%)                           &                                                   \\
                                      & ALIGNet \cite{hanocka2018alignet}   & Local Shape-dependent Chamfer Upperbound & 31.37 (62\%)                           &                                                   \\
                                      & ProAlignNet                                          & Normalized Cross-Correlation             & 32.78 (61\%)                           &                                                   \\
                                      & ProAlignNet                                          & Asymmetric Chamfer                       & 28.05 (67\%)                           &                                                   \\
                                      & ProAlignNet                                          & Local Shape-dependent Chamfer Upperbound & \textbf{20.63 (75\%)} &                                                   \\ \hline
\end{tabular}
    \caption{\small Quantitative evaluations: Chamfer scores are reported along with percentage (in the brackets) of pixels whose misalignment is less than $Z$ pixels. We use $Z=5$ for MNIST shape alignment and $Z=20$ for geo-parcel alignment.}
    \label{tab_quantitative}
\vspace{-15pt}
\end{table*}

\section{Ablation Studies using contourMNIST} \label{exp_mnist}
We first evaluate the behavior of our proposals on a simulated dataset. Several recent works \cite{eslami2016attend,deng2018probabilistic} adopted MNIST dataset \cite{deng2012mnist} to simulate toy datasets to understand the behavior of the models and training processes without requiring lengthy training periods. Following this trend we also simulate a dataset adopting MNIST digits to the current context of noisy contour alignment. \\
\textbf{Simulating contourMNIST dataset}: MNIST database contains 70000 gray-scale images of handwritten digits of resolution $28\times28$. The test images (10,000 digits) were kept separate from the training images (60,000 digits). In our experiments, each digit image is upsampled to $128\times128$ resolution and converted as a contour image. Each contour image is then transformed with randomly generated \textit{tps} transformations to create a misaligned source contour image while the original one considered as the target image.  We add noise and occlusions randomly to simulate partial noisy observations of the contours, as shown in Figure \ref{fig_mnist}. Now the task considered in this section is to align these noisy source to target shapes. 

\begin{figure*}[ht!]
    \centering
    \subfloat[Aerial image\label{fig_parcel_align_aerial_img} ]{
    \begin{tabular}{c}
        \includegraphics[width=0.22\textwidth]{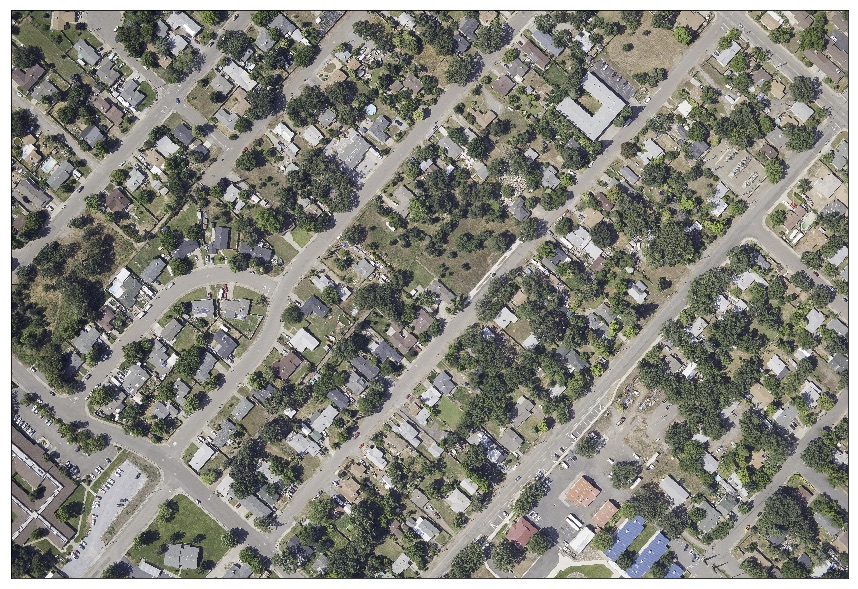}\\
        \includegraphics[width=0.22\textwidth]{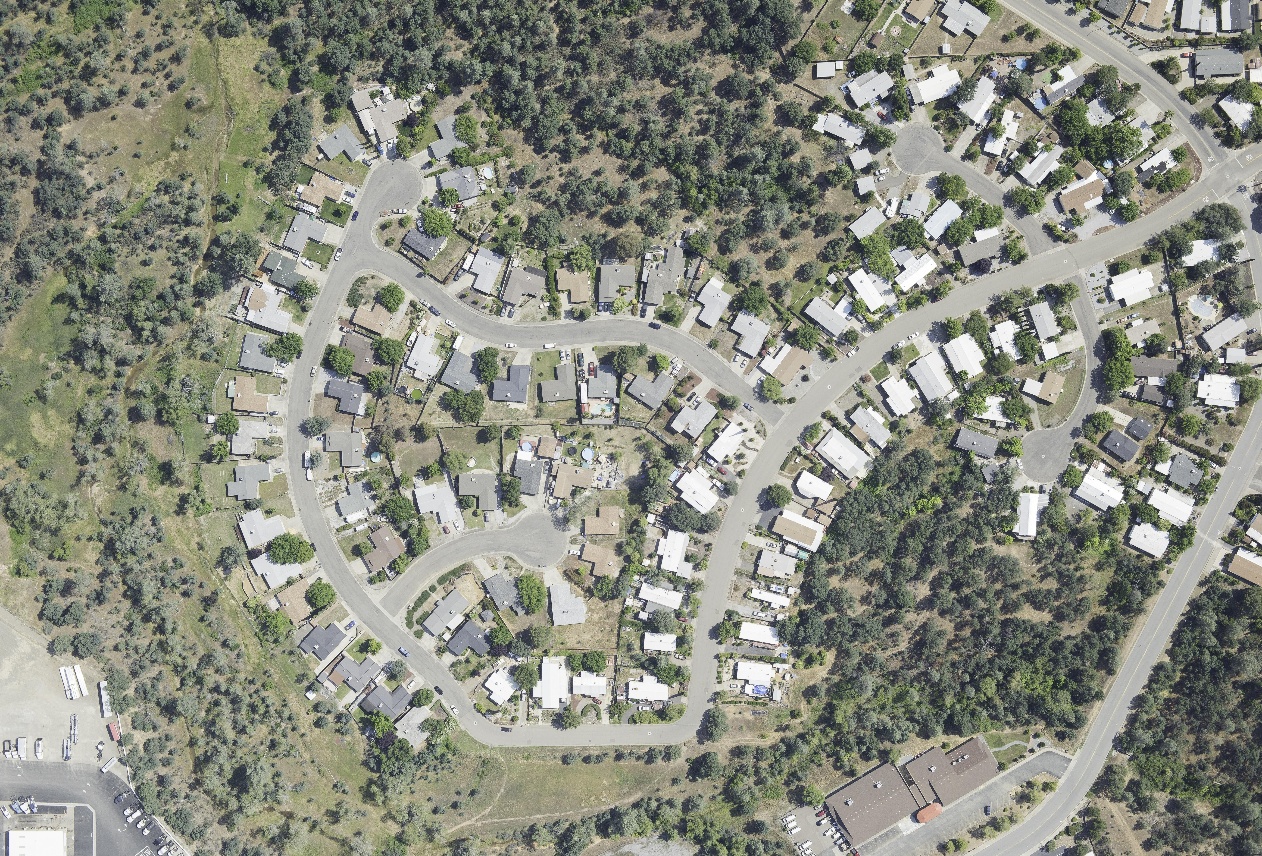}  
    \end{tabular}
    }
    \subfloat[Road contours from (a)\label{fig_parcel_align_contour_img}]{
        \begin{tabular}{c}
            \includegraphics[width=0.22\textwidth]{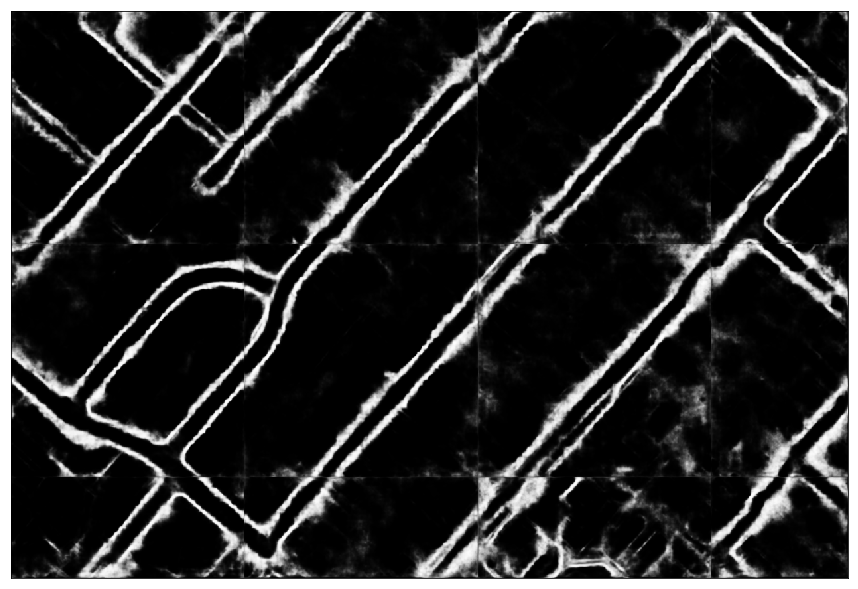} \\
            \includegraphics[width=0.22\textwidth]{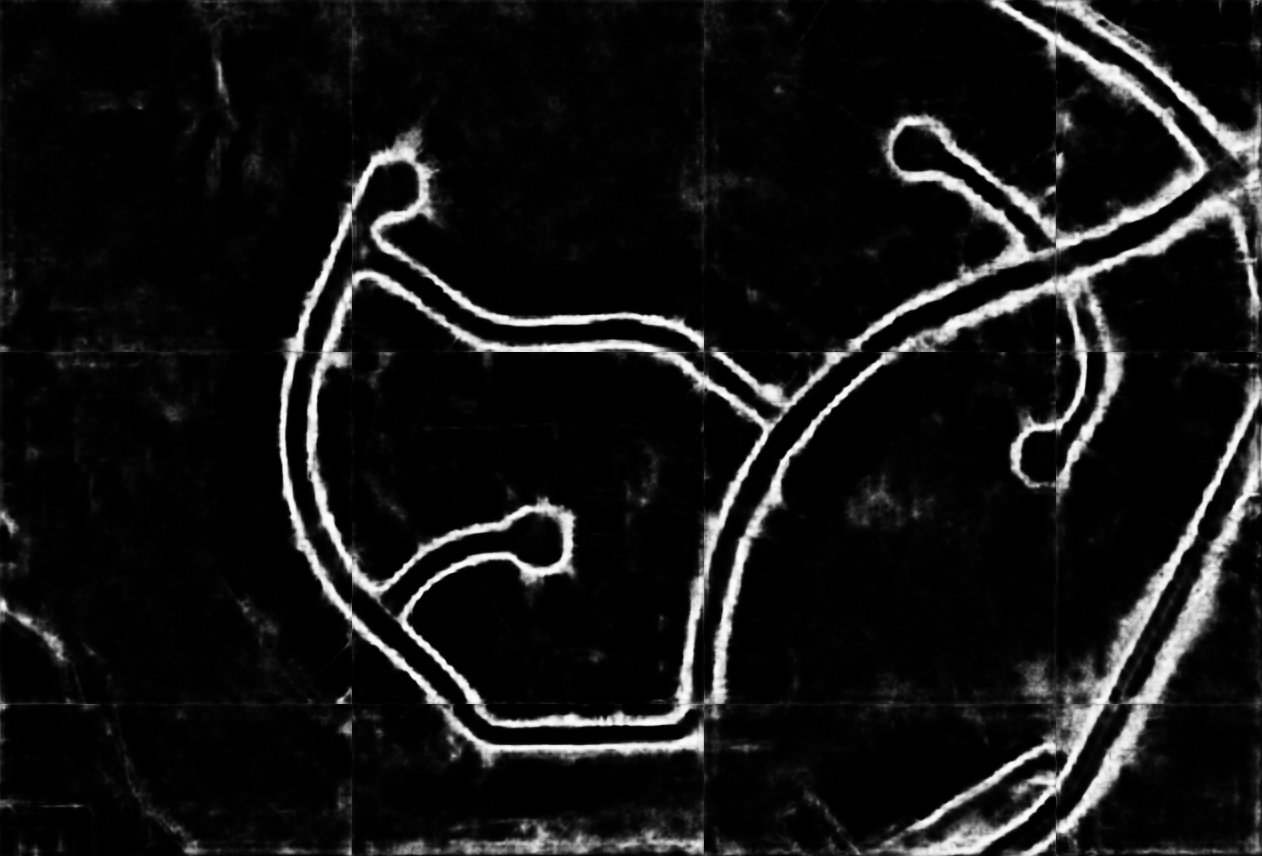}
        \end{tabular}
    }
    \subfloat[Geo-parcels  \label{fig_parcel_align_parcel_img}]{
    \begin{tabular}{c}
        \includegraphics[width=0.22\textwidth]{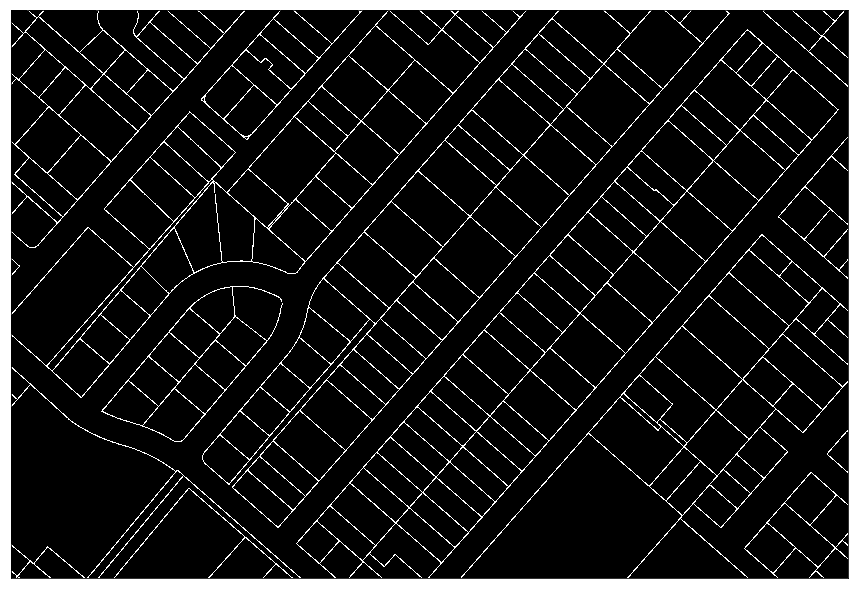} \\
        \includegraphics[width=0.22\textwidth]{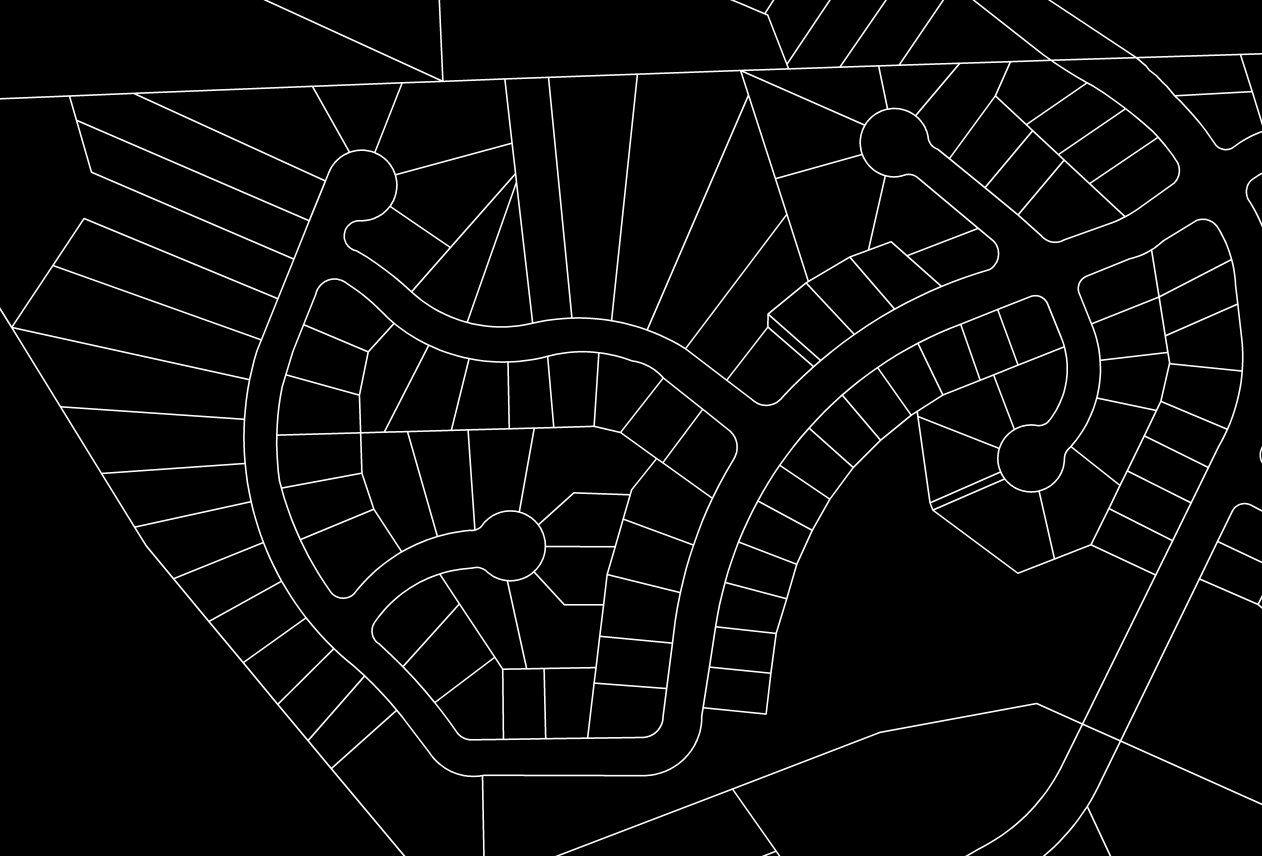}
    \end{tabular}
    }
    \subfloat[Aligned parcels overlaid on (a) \label{fig_parcel_align_overlaid}]{
    \begin{tabular}{c}
         \includegraphics[width=0.22\textwidth]{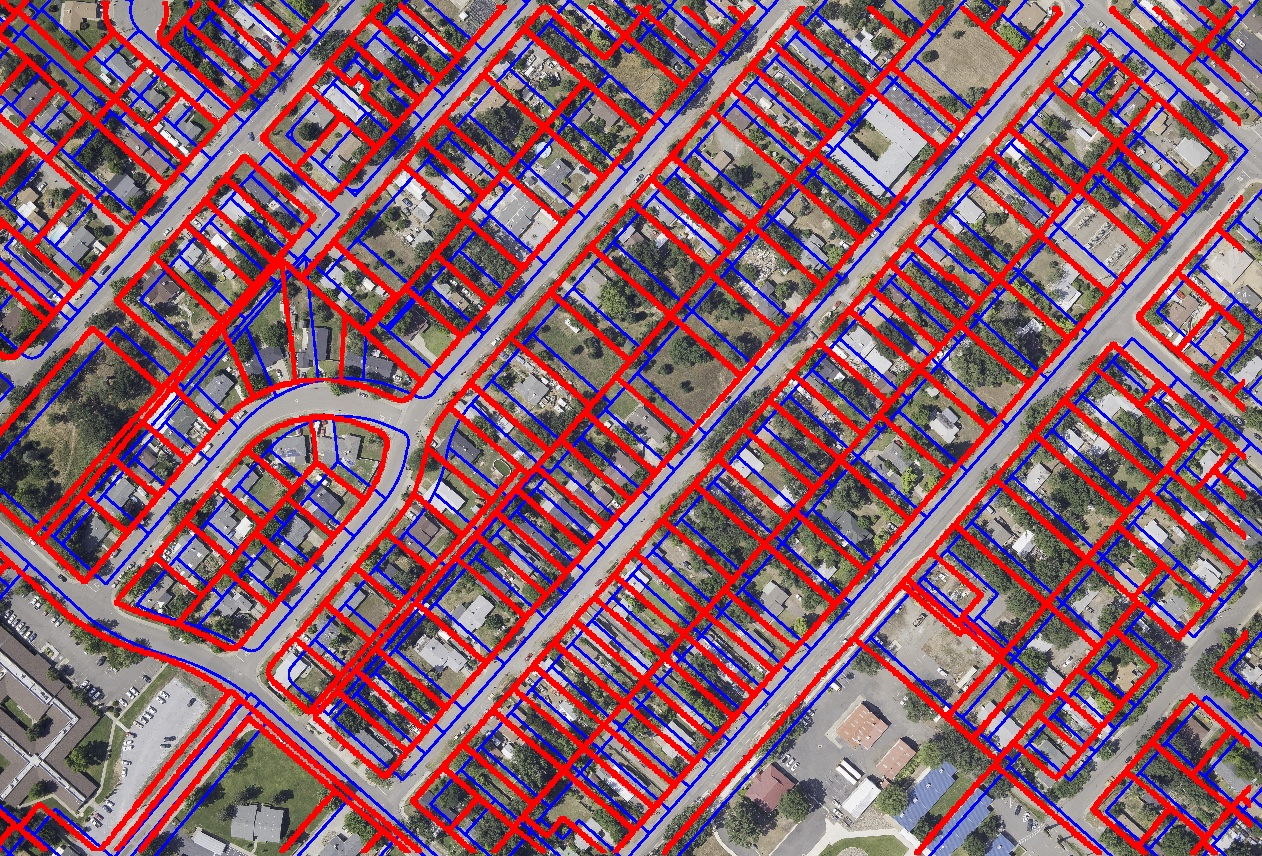} \\
         \includegraphics[width=0.22\textwidth]{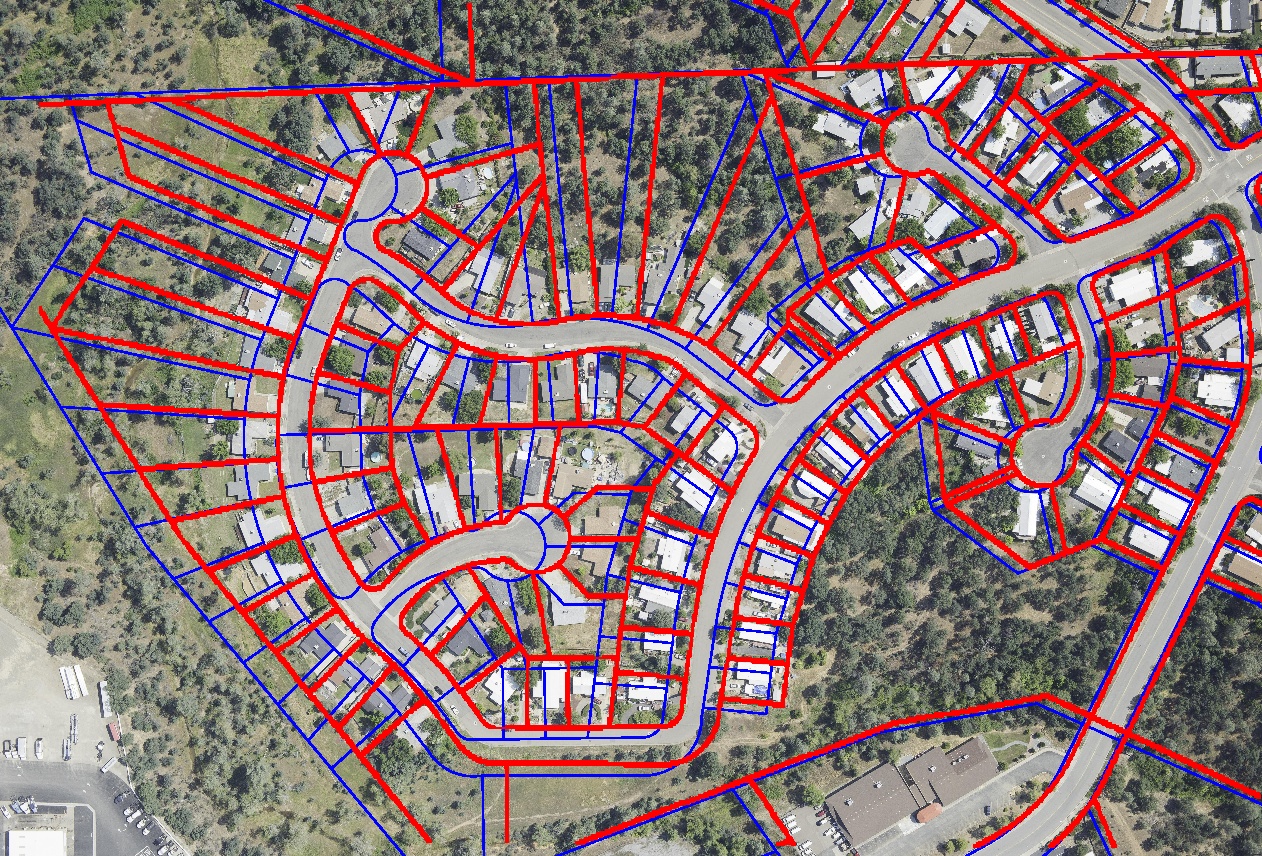}
    \end{tabular}
    }
    \caption{Geo-parcel alignment with road evidences. Better see in enlarged version.}
    \label{fig_parcel_align}
    \vspace{-15pt}
\end{figure*}

\textbf{Implementation details}: To implement the ProAlignNet architecture for this task, we use CNN bases with five convolutional blocks. Each block is composed with three conv (with filter size $5\times5$) + leakyReLU (with slope 0.1) layers followed by a MaxPool (with stride 2 and kernel size $2\times2$) layer. 
Multiscale alignment part consists of five warp predictors ($P^{(i)}$) that operate on five scales and predict increasingly complex transforms. In this application, we use affine transforms at the coarsest scale and \textit{tps} with control grids of increasing resolutions ($2\times2$, $4\times4$, $8\times8$, $16\times16$) at finer scales. The warp predictor blocks are composed of three conv layers and a fully connected MLP. Each conv layer is followed by leakyRelu + MaxPool except the last one. MLP takes in the flatten conv features from the last layer and predict parameters required for the transform at that scale. For instance, \textit{tps} warp predictor outputs $2n+6$ parameters while affine predictor output $6$ parameter values. We set $\lambda_i=1.0$ and $\alpha=1e-2$ for all the experiments in this paper.\\
\textbf{Training}: We train the networks for 10 epochs in a completely unsupervised manner using the losses mentioned in Section \ref{sec_losses}.  We use SGD optimizer with learning rate as $1e-5$.\\
\textbf{Evaluation metric}: We use asymmetric Chamfer distance b/w aligned source and target images (non-noisy versions), $\frac{1}{N} S(\theta_{S\rightarrow T}).dt[T]$, as it measures average misalignment in the units of pixels. We also measure the percentage of pixels whose misalignment under 5 pixels. As shown in Table \ref{tab_quantitative}, test set image pairs are misaligned by an average of 10.20 pixels. The percentage of pixels with misalignment under 5 pixels is 39\%. 

We now evaluate the different aspects of our proposals via a set of experiments. \\
\textbf{ProAlignNet vs Baselines} : Here we start by comparing the performance of ProAlignNet with that of two baselines, DIRNet \cite{de2017end} and ALIGNet \cite{hanocka2018alignet}, when trained with different loss functions, including the ones used in their implementations and our Chamfer loss functions.
 Results in Table \ref{tab_quantitative} show that ProAlignNet outperforms DIRNet and ALIGNet in all experimental settings. The best performance is achieved with ProAlignNet model trained using Chamfer uppperbound loss. This is atleast 25\% better than the models of DIRNet or ALIGNet. We believe that this superior performance is due to multiscale inference and progressive alignment processes in our network architecture.  \\
\textbf{Chamfer upperbound vs Other loss functions}: The test performances of both DIRNet and ALIGNet has been boosted up by 23\% and 16\% respectively when trained with chamfer upperbound loss rather than the ones (NCC, MSE) used in their implementations. \\
\textbf{Bidirectional vs Asymmetric Chamfer loss}: ProAlignNet yields 4\% better performance when trained with bi-directional reparametrized Chamfer loss compared to asymmetric Chamfer loss.  This improvement demonstrates the regularization capabilities of forward-backward consistency constraints in the loss. Overall, the best performance of 96\% is achieved when ProAlignNet is trained with shape-dependent Chamfer upperbound. This is approximately 7\% better than the model that is trained with asymmetric loss.\\
\textbf{Impact of local shape dependency}: Similarly, the model trained by shape-dependent Chamfer loss performs better than the one trained with bi-directional Chamfer loss by approximately 3\%. 
\vspace{-5pt}

\section{Geo-parcel to Aerial Image Alignment} \label{exp_geo}
\textbf{Problem statement}: In this section, we discuss aligning geo-parcel data to aerial images. Geo-parcel data is used to identify public and private land property boundaries. Parcels are shapefiles with lat-long gps coordinates of the property boundaries maintained by local counties.  One can project these parcel shapes (using perspective projection) onto the coordinate system of the camera with which aerial imagery was captured. This process results in binary contour images as shown in Figure \ref{fig_parcel_align_parcel_img}. These contours are ideally expected to match visual contours in the aerial image of the corresponding region (shown in Figure \ref{fig_parcel_align_aerial_img}). However, due to several differences in their collection processes, these two modalities of the data often misalign by a large extent, sometimes, in the order of 10 meters. Figure \ref{fig_parcel_align_overlaid} depicts the misalignment of the original (before alignment) parcel contours overlaid on the aerial image in blue color. In this work, we extract the road (including sidewalks) contours from aerial images using an off-the-shelf contour detection method \cite{yu2017casenet} and consider these as target contours. These extracted contours are noisy and partial in shape, as seen in Figure \ref{fig_parcel_align_contour_img}. We train our ProAlignNet to align parcel contours with these road contours from aerial imagery. 

\textbf{Dataset}: Our dataset contains 1189 aerial and parcel image pairs captured over residential areas of Redding, California. Fortunately, our method does not require any ground-truth alignment parameters to train. However, we prepare a validation set for which we manually aligned 27 parcel-aerial image pairs with more than 7000 parcel polygons. \\
\textbf{Implementation details}: Experimental settings are similar to the above section. However, we work with input resolution of $1024\times 512$ resolution for this application. 

\textbf{Evaluation metric}: In addition to average misalignment (Chamfer score), we also report the percentage of pixels with misalignment under 3ft (20 pixels as ground-sampling-distance is 4.5cm/px for our aerial data). 

\textbf{Evaluation}:
Parcel data aligned with ProAlignNet (trained with Local Shape-dependent Chamfer Upperbound) is overlaid with red color in Figure \ref{fig_parcel_align_overlaid}. As one can see, it is aligned well with the aerial image contents than the original parcels (blue color). Results in Table \ref{tab_quantitative} show that ProAlignNet outperforms both DIRNet and ALIGNet even when trained with NCC. 
Best performance (75\%) is achieved when ProAlignNet is trained with Local Shape-dependent Chamfer Upperbound.
Alignment quality of these parcels with the corresponding aerial image contents has been improved by 27\% with ProAlignNet. Moreover, training with our Chamfer upperbound loss has boosted up the performances of DIRNet and ALIGNet, similar to the behavior observed on contourMNIST data. 
\vspace{-7pt}

\section{Refining Coarser Segmentation Annotations}
\begin{figure}[ht!]
    \centering
    \includegraphics[width=0.23\textwidth]{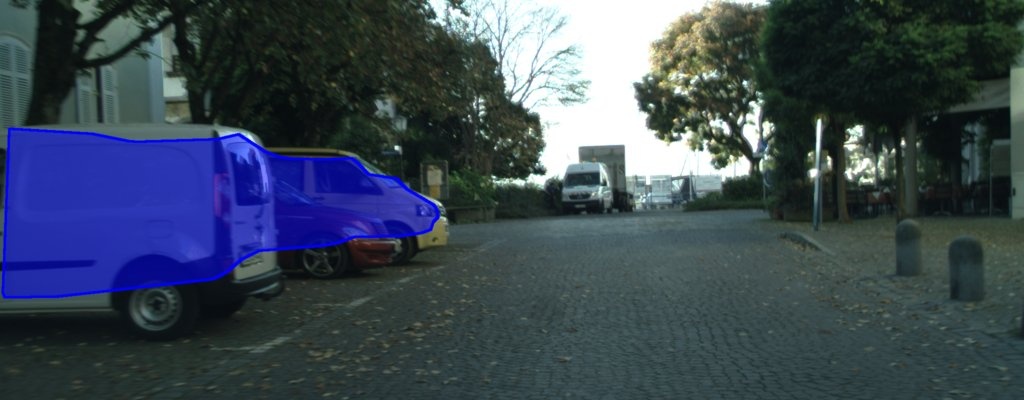}
    \includegraphics[width=0.23\textwidth]{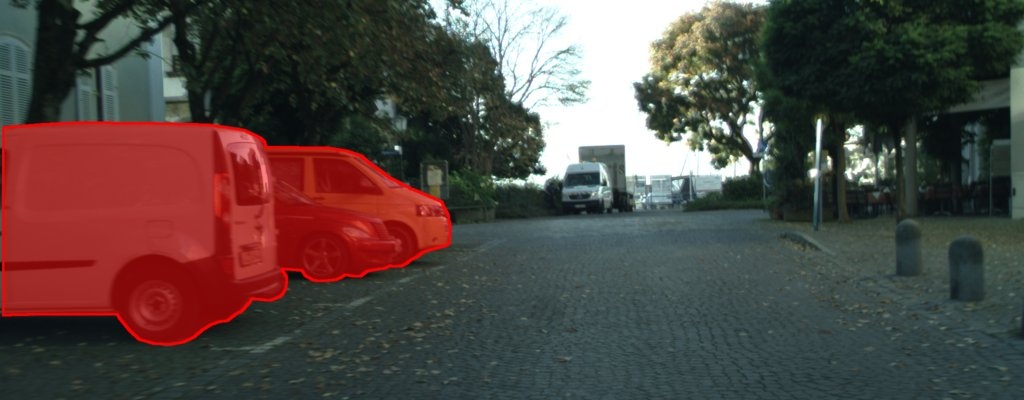}
    \caption{\small Refining Coarser Labels: Coarser annotations (blue parts in left image) are refined to extract precise labels (red parts in right image).}
    \label{fig_cs_coarse_2_refine}
    \vspace{-15pt}
\end{figure}

In this section we demonstrate that how the proposed models can be used to refine coarsely annotated segmentation labels. 
For these experiments we use CityScapes dataset \cite{cordts2015cityscapes}, a publicly available benchmark for traffic scene semantic segmentation. It provides the data in 3 sets with public access to GT labels: \textit{train} (2975 samples), \textit{val} (500 samples), and \textit{train-extra} (19998 samples) sets. The image samples in the larger set, \textit{train-extra}, have only coarser annotations (See Figure \ref{fig_cs_coarse_2_refine} \& \ref{fig_intro}), while the sets of \textit{train} and \textit{val} provide both finely and coarsely annotated GT labels. Refining the coarser annotations using ProAlignNet is our task of interest in this section. 
We use an off-the-shelf semantic contour extraction method, CASENet \cite{yu2017casenet} (pretrained on \textit{train} set) to get contours from the images which are treated as target shapes in the current context. The given coarsely annotated labels are considered as source contours and to be aligned with contour predictions from the CASENet model. Here, we train our ProAlignNet using the chamfer upperbound loss on \textit{train} set. 
Experimental set up is similar to above sections except that tps control grid resolutions in $P^{(3)}$ and $P^{(4)}$ (see Figure \ref{fig_pipeline}) have been doubled for this application.
For quantitative analysis, we use \textit{val} set for which both coarse and fine labels are available. Qualitative results are shown on \textit{train-extra} images. 
\begin{table}[ht!]
    \tiny
    \centering
    \begin{tabular}{|c|c|c|c|c|c|} 
    \hline
    Label Quality & 4px error & 8px error & 16px error & 32px error & Real Coarse  \\ \hline
    Num Clicks per Image & 175.23 & 95.63 & 49.21 & 27.00 & 98.78 \\ 
    Test IoU & 74.85 & 53.32 & 33.71 & 19.44 & 48.67 \\ \hline \hline
    GrabCut \cite{rother2004grabcut}  & 26.00 & 28.51 & 29.35 & 25.99 & 32.11 \\ \hline
    STEALNet \cite{acuna2019devil} & 78.93 & 69.21 & 58.96 & 50.35 & 67.43 \\ \hline
    ProAlignNet (Ours)  & 79.41 & 69.73 & 67.51 & 61.05 & 71.45 \\ \hline
    \end{tabular}
    \caption{\small Model trained on \textit{train} set and used to refine coarse data on \textit{val} set. Real Coarse corresponds  to coarsely human annotated val set, while x-px error correspond to simulated coarse data. Score (\%) represents mean IoU.}
    \label{tab_noise_impact_coarse2fine}
    \vspace{-10pt}
\end{table}

\textbf{Coarse Label Simulation}: For training data augmentation and quantitative study (shown in Table \ref{tab_noise_impact_coarse2fine}), we also synthetically coarsen the given finer labels following the procedure described in \cite{acuna2019devil,zlateski2018importance}. This synthetic coarsening process first erodes the finer segmentation mask and then simplifies mask boundaries using Douglas-Peucker polygon approximation method to produce masks with controlled quality. 
Intersection-over-Union (IoU) metrics b/w these coarser and finer labels of \textit{val} set are shown in Table \ref{tab_noise_impact_coarse2fine}. 
We also count the number of vertices in the simplified polygon and report it in Table \ref{tab_noise_impact_coarse2fine} as an estimate of the number of clicks required to annotate such object labels. 

\begin{figure}
    \centering
    \includegraphics[width=0.48\textwidth, height=4.5cm]{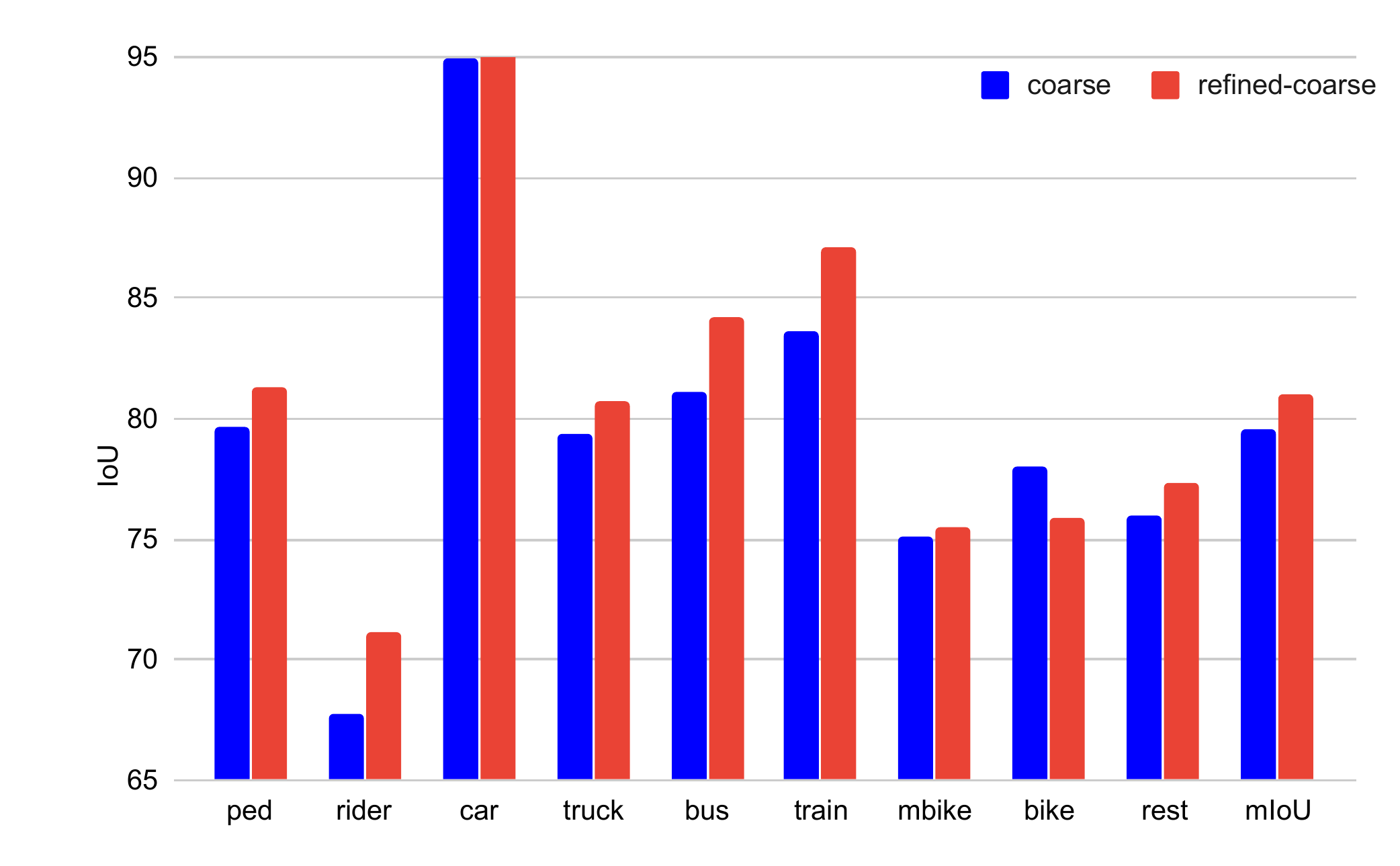}
    \caption{\small Semantic Segmentation on Cityscapes \textit{val} set: Performance of UNet when trained with (in addition to \textit{train} set) coarse labels vs our refined labels of \textit{train-extra} set. We see improvement of more than 3 IoU \% in rider, bus and train.}
    \label{fig_segment}
    \vspace{-15pt}
\end{figure}

\textbf{Results}: A recent work, STEALNet \cite{acuna2019devil}, addressed this problem of refining coarser annotation labels. Hence, we use it as a baseline for comparison along with GrabCut tool \cite{rother2004grabcut}. As reported in Table \ref{tab_noise_impact_coarse2fine}, our method perform equally well with STEALNet at lower-scale misalignments. However, the superiority of our ProAlignNet is quite evident with larger misalignments (16px, 32px errors). Also, our method is better by $\sim 4$\% compared to STEALNet on real coarser labels of \textit{val} set,.
As shown in Figure \ref{fig_cs_coarse_2_refine}, by starting from a very coarse segmentation mask our method is able to obtain very precise refined masks. Hence, we think that our approach can be introduced in current annotation tools saving considerable amount of annotation time.

\textbf{Improved Segmentation}:  We also evaluate whether our refined label data is truly beneficial for training segmentation methods. Towards this end, we refine 8 object classes in the whole \textit{train-extra} set. We then train our implementation of UNet based semantic segmentation architecture \cite{ronneberger2015u} with the same set of hyper-parameters with and without refinement on the coarse labels of \textit{train-extra} set. Individual performances (IoU\%) on the 8 classes are reported in Figure \ref{fig_segment}.  Training with refined labels results in improvements of more than 3 IoU\% for rider, bus and train as well as 1.5 IoU\% in the overall mean IoU (79.52 vs 81.01).

\vspace{-5pt}
\section{Conclusions}
\vspace{-5pt}
This work introduced a novel ConvNet-based architecture, ``\textit{ProAlignNet}," that learns --without supervision-- to align noisy contours in multiscale fashion by employing progressively increasing complex transformations over increasing finer scales. We also proposed a novel proximity-measuring and local shape-dependent Chamfer distance based loss. 
The sanity checks for behaviors of the proposed networks and loss functions have been done using a simulated contourMNIST dataset. 
We also demonstrated the efficacy of the proposals in two real-world applications: (a) aligning geo-parcel data with aerial imagery, (b) refining coarsely annotated segmentation labels. In future, we explore more effective feature fusing schemes for warp predictors and sophisticated/learnable shape metrics.\\
\small{\textbf{Acknowledgments:} We thank GEOMNI (www.geomni.com) for providing the data for geo-parcel alignment application.}


{\small
\bibliographystyle{ieee_fullname}
\bibliography{ref}
}

\end{document}


\title{Supplementary Material for \\ ProAlignNet : Unsupervised Learning for Progressively Aligning Noisy Contours}

\author{VSR Veeravasarapu, Abhishek Goel, Deepak Mittal, Maneesh Singh\\
Verisk AI, Verisk Analytics \\
{ \tt\small \{s.veeravasarapu, a.goel, deepak.mittal, msingh\}@verisk.com }
}

\maketitle

\section*{A1: Proof for $dt[S(\theta_{S\rightarrow T})].T = dt[S].T(\theta_{T\rightarrow S})$}
Under the assumption of infinite boundaryless coordinate systems, 
$\mathbb{S}$ is source coordinate system. 
$\theta_{S\rightarrow T}$ is a transformation require to align $\mathbb{S}$ with target coordinate system: $\mathbb{T} \equiv  \mathbb{S}(\theta_{S\rightarrow T}) $

\textbf{Homeomorphism}: Most of the transformations we use for contour alignment applications are homeomorphic. The property of homeomorphism states that for each transformation that aligns S to T, there exists an inverse/backward transform that aligns T to S. 
\begin{equation}
    if \ S(\theta_{S\rightarrow T}) \equiv T ,  \ \ \ \exists \ \ \theta_{T \rightarrow S} \in \Theta \ s.t.\ T(\theta_{T\rightarrow S}) \equiv S
\end{equation}

s and t represent all nonzero pixel locations in S and T respectively.
\begin{equation}
\small
\begin{split}
    dt[S(\theta_{S \rightarrow T})].T & = \sum_{t \in  \mathbb{T}} \min_{s\in \mathbb{S}(\theta_{S\rightarrow T})} \|t-s\|^2 \\
    & change-of-coordinates \\
    & =   \sum_{t \in  \mathbb{T(\theta_{T \rightarrow S})}} \min_{s\in \mathbb{S}} \|t-s\|^2 \\
    & = dt[S].T(\theta_{T \rightarrow S})
\end{split}
\end{equation}

\section*{A2: Min-Max Inequality}

\textbf{Proof for $\min_x\big(f(x)+g(x)\big) \le \min_x f(x) + \max_x g(x)$
}

Let us start with below inequality that holds for any $x$, 

\begin{equation}
  \big( f(x)+g(x)\big) \le \big(f(x) + \max_x g(x) \big)
\end{equation}

The above inequality holds for even smallest possible values of both LHS and RHS. 
\begin{equation}
    \min_x\big(f(x)+g(x)\big) \le \min_x \big( f(x) + \max_x g(x) \big)
\end{equation}
$\max_x g(x)$ is constant wrt $x$. 
\begin{equation}
    \min_x\big(f(x)+g(x)\big) \le \min_x f(x) + \max_x g(x)
\end{equation}

\section*{A3: Detailed derivation for Local shape dependent Chamfer Upperbound}
We use min-max inequality to reformulate Eq 5 (in the main paper) so that one can use the concepts of MDT and reparameterization as in Eq 3 and 4 (in the main paper). Min-max inequality states that minimum of the sum of any two arbitrary functions $f(x)$ and $g(x)$ is upper-bounded by the sum of minimum and maximum of individual functions. (Please refer to the appendix A2 for the proofs). 
\begin{equation}
    \min_x\big(f(x)+g(x)\big) \le \min_x f(x) + \max_x g(x)
\end{equation}
Under mild assumptions ($\max_x f \ge \max_x g$), one can prove that this is the tightest possible upperbound. Using the above inequality for the first term on RHS (right hand side) of Eq 5 results in, 
\begin{equation}
\begin{split}
   \sum_{x \in X} \min_{y \in Y}\big( E(x,y) & + \lambda E(I_x', I_y') \big)  \\
   & \le   \sum_{x \in X}\min_{y \in Y}  E(x,y) + \lambda \sum_{x \in X}\max_{y \in Y} E(I_x', I_y')  
    \end{split}
\end{equation}

Using the inequality for both terms on RHS of Eq 5 results in an upperbound with original Chamfer distance (Eq 2) and shape-dependent terms as follows,
\begin{equation}
\begin{split}
& C_d(X,Y) \le   C(X,Y) \\
& + \lambda \bigg( \frac{1}{N_X} \sum_{x \in X}\max_{y \in Y} E(I_x', I_y') + \frac{1}{N_Y} \sum_{y \in Y}\max_{x \in X} E(I_y', I_x')  \bigg)  
\end{split}
\end{equation}

Rewriting the above upperbound in the current context of source to target alignment,  
\begin{equation}
\begin{split}
& C_d(S(\theta),T) \le   C(S(\theta),T) \\
& + \lambda \bigg( \frac{1}{N_{S(\theta)}} \sum_{x \in S(\theta)}\max_{y \in T} E(I_x', I_y') + \frac{1}{N_T} \sum_{y \in T}\max_{x \in S(\theta)} E(I_y', I_x')  \bigg)  
\end{split}
\end{equation}
We denote this upperbound as $C_{up}$. As one can observe, the shape-dependent terms are computationally heavy as the maximum being taken over the window of the entire image for each pixel in the other image. However, we can constraint this window to be local and search in the neighborhood defined by that window. Moreover, this maximum-finding operation can be implemented with \textit{MaxPool} layers. Finally, this local shape-dependent Chamfer upperbound is given by, 
\begin{equation}
\begin{split}
    C_{up}(S, T) = & \bigg( dt[S].T(\theta_{T\rightarrow S}) + S(\theta_{S\rightarrow T}).dt[T] \bigg) + \\ 
    & \lambda \bigg(\sum_{x\in S(\theta)} \max_{{y\in T_x}} E(I_x',I_y')   + \sum_{y\in T} \max_{{x\in S_y(\theta)}} E(I_y',I_x')\bigg)
\end{split}
\end{equation}




As mentioned, this upperbound is a weighted combination of Chamfer loss that measures proximity and a local shape-dependent loss. It is interesting take a closer look at the shape-dependent terms which are distances between two unit gradients. 
\begin{equation}
\begin{split}
    E(I_x',I_y') & = \sqrt{|I_x'- I_y'|_2^2} \\
    & = \sqrt{I_x'^2+I_y'^2-2I_x'^T.I_y'} \\
    & \propto \sqrt{1-I_x'^T.I_y'}   
\end{split}
\end{equation}
When the local window is restricted to be $1\times1$, minimizing the above term is related to maximizing cross-correlation in intensity gradient space. When raw pixel intensities are used in place of gradients, this is maximizing NCC-related metric.

Now the upperbound loss with unit gradients as local shape measures, 
\begin{equation}
\begin{split}
    C_{up}\big(S(\theta), T\big) = & \bigg( dt[S].T(\theta_{T\rightarrow S}) + S(\theta_{S\rightarrow T}).dt[T] \bigg) \\
    &+ \lambda \bigg(  \sum_{x\in S(\theta)} \max_{{y\in T_x}} \sqrt{1-I_x'^T.I_y'} \\ 
    &+ \sum_{y\in T} \max_{{x\in S_x(\theta)}} \sqrt{1-I_y'^T.I_x'}\bigg)
\end{split}
\end{equation}




\section*{A4: Background for Multiscale Feature based Approaches}
Our designs in this work are majorly inspired by several principles followed in the classical literature. One of them is using multiscale features for incremental alignment. 
In the classical literature of image registration and optical flow fields, large scale displacements are addressed with multiscale approaches \cite{saxena2014survey,klein2009elastix,sun2010secrets}. These schemes help to increase search scope and escape local minima \cite{lindeberg1994scale}.

These multiscale problem-solving practices have recently regained attention in deep network architectures. Several recent approaches use input images or their features at different resolutions to improve object detection \cite{lin2017feature} and segmentation \cite{ronneberger2015u,pan2017fully}. 
Recent versions of several popular object detection frameworks such as 
YOLO \cite{redmon2017yolo9000} employed multiscale processes to detect objects at multiple scales of the feature hierarchy independently and fuse them for robust detections. 
For semantic label and contour estimation, several works \cite{ronneberger2015u,yu2017casenet} exploit lateral/skip connections that associate low-level feature maps across resolutions and semantic levels.
Inspired by the effectiveness of multiscale processes in classical and modern literature, we design a deep network that solves alignment problem incrementally at multiple scales. 

\section*{A5: Background for Progressive Transformations }
In classical literature \cite{lowe1999object,philbin2007object} another commonly used practice when estimating complex transformations is to start by estimating a simple transform such as affine and then progressively increase the transform-complexity to refine the estimates along the way. The motivation behind this practice is that estimating a very complex transformation could be hard and computationally inefficient in the presence of noise, so a robust and fast rough estimate of a simpler transformation can be used as a starting point, also regularizing the subsequent estimations of the more complex transformations.

In this work, we follow this practice as we deal with complex misalignments. We start at coarser scales by estimating an affine transformation, which is a linear transformation with 6 degrees-of-freedom (DOF), capable of modeling translation, rotation, non-isotropic scaling, and shear. 
This estimated affine grid is refined through finer scale networks which employ more flexible transformations (for example, thin-plate splines in this work). 



{\small
\bibliographystyle{ieee_fullname}
\bibliography{supple_ref}
}